\documentclass{article}

\PassOptionsToPackage{authoryear,round}{natbib}
\usepackage{graphicx}
\usepackage{amsmath}
\usepackage{adjustbox} 
\usepackage{tikz}
\usepackage{pgfplots}
\pgfplotsset{compat=1.18}
\usepgfplotslibrary{polar}

 \usepackage[main, final]{neurips_2025}

\usepackage[utf8]{inputenc} 
\usepackage[T1]{fontenc}    
\usepackage{hyperref}       
\usepackage{url}            
\usepackage{booktabs}       
\usepackage{amsfonts}       
\usepackage{nicefrac}       
\usepackage{microtype}      
\usepackage{xcolor}         

\title{Dual-Head Reasoning Distillation: Improving Classifier Accuracy with Train-Time-Only Reasoning}

\newif\ifcomments
\commentstrue
\ifcomments
    \newcommand{\vinay}[1]{\textcolor{orange}{~Vinay:~#1}}
    \newcommand{\dylan}[1]{\textcolor{blue}{~Dylan:~#1}}
    \newcommand{\shuhuai}[1]{\textcolor{purple}{~Shuhuai:~#1}}
    \newcommand{\jillian}[1]{\textcolor{red}{~Jillian:~#1}}
\else
    \newcommand{\vinay}[1]{}
    \newcommand{\dylan}[1]{}
    \newcommand{\shuhuai}[1]{}
    \newcommand{\jillian}[1]{}

\fi

\author{%
  Jillian Xu\thanks{Work performed while working at Google} \\
  University of Waterloo\\
  \texttt{j23xu@uwaterloo.ca} \\
  \And
   Dylan Zhou \\
  Google\\
  \texttt{dylanzhou@google.com} \\
   \And
   Vinay Shukla \\
  Google\\
  \texttt{vinayshukla@google.com} \\
   \AND
   Yang Yang \\
  Google\\
  \texttt{lizyang@google.com} \\
  \And
   Junrui Ruan \\
  Google\\
  \texttt{junrui@google.com} \\
  \And
   Shuhuai Lin \\
  Google\\
  \texttt{shuhuailin@google.com} \\
   \AND
   Wenfei Zou \\
  Google\\
  \texttt{wenfei@google.com} \\
   \And
   Yinxiao Liu \\
  Google DeepMind\\
  \texttt{canoee@google.com} \\
  \And
   Karthik Lakshmanan \\
  Google \\
  \texttt{lakshmanan@google.com} \\
}

\begin{document}

\maketitle

\begin{abstract}
     \textit{Chain-of-Thought} (CoT) prompting often improves classification accuracy but it introduces a significant throughput penalty with rationale generation  \citep{Wei2022CoT,Cheng2024CCoT}. To resolve this trade-off, we introduce \emph{Dual-Head Reasoning Distillation} (DHRD), a simple training method for decoder-only language models (LMs) that adds (i) a pooled \emph{classification head} used during training and inference and (ii) a \emph{reasoning head} supervised by teacher rationales used only in training. We train with a loss function that is a weighted sum of label cross-entropy and token-level LM loss over input-plus-rationale sequences. On seven SuperGLUE tasks, DHRD yields relative gains of $0.65$--$5.47\%$  over pooled baselines, with notably larger gains on entailment/causal tasks. Since we disable the reasoning head at test time, inference throughput matches pooled classifiers and exceeds CoT decoding on the same backbones by 96--142$\times$ in QPS. 
\end{abstract}
\section{Introduction}
\label{sec:intro}

Decoder-only LMs are widely adapted to encoder-style classification tasks by pooling hidden states and applying a lightweight classifier (e.g. \texttt{AutoModelForSequenceClassification} from HuggingFace \citep{Wolf2020Transformers}).  While effective, this adaptation can under-utilize the model’s latent reasoning abilities learned in pre-training. CoT can surface that capacity, but its token-by-token decoding can significantly reduce throughput, making it impractical for high-throughput applications \citep{Wei2022CoT,Cheng2024CCoT}.

\begin{figure}[htpb]
    \centering
    \includegraphics[width=1\linewidth]{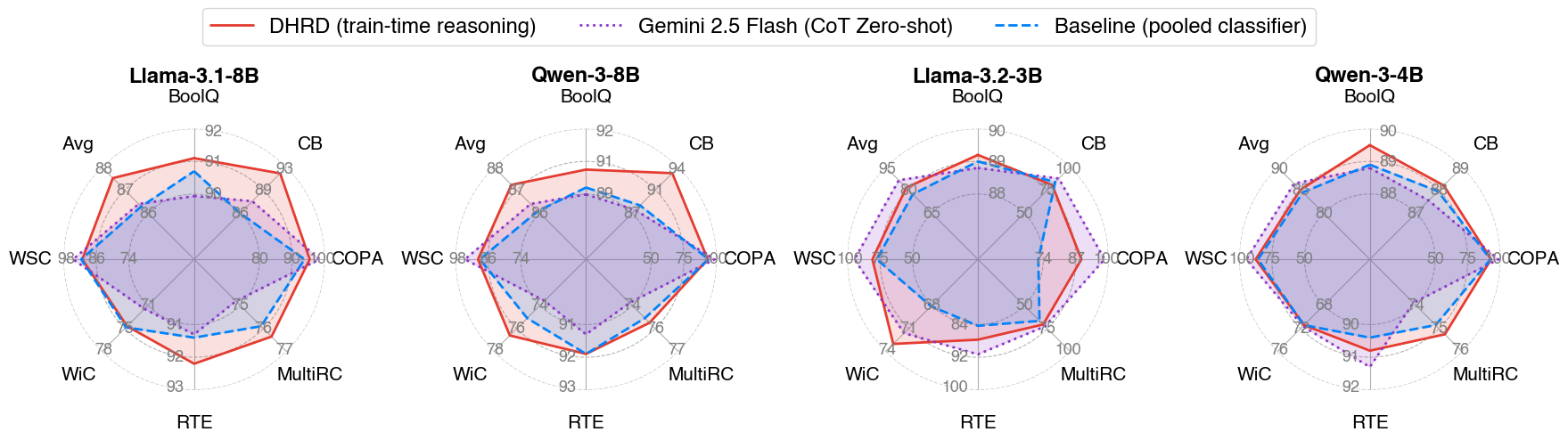}
    \caption{SuperGLUE per-task scores for four backbones. DHRD (train-time reasoning) consistently beats the pooled-classifier baseline and rivals teacher model \textit{Gemini 2.5 Flash}, with the largest gains on CB/COPA/RTE. ‘Avg’ is the macro-average, tabulated results can be found in Table~\ref{tab:sg-eval}.}
    \label{fig:spiderwebs}
\end{figure}

We propose \emph{Dual-Head Reasoning Distillation} (DHRD), a simple method that requires just two small additions to the backbone decoder: (i) a classification head that pools token representations with last-token pooling, and (ii) a reasoning tower/LM head used only for auxiliary training loss. We optimize a weighted joint objective that balances label cross-entropy and auxiliary token-level cross-entropy; a CoT-capable teacher (we use \emph{Gemini 2.5 Flash} (\citealp{Gemini25Report2025})) prompted with Zero-shot Chain-of-Thought (‘Let’s think step by step’; \citep{Kojima2022ZeroShotNeurIPS}) provides rationales for training (for prompt refer to Appendix ~\ref{app:teacher}). We evaluate on SuperGLUE tasks under standard model input formatting and metrics (refer to Appendix \ref{app:fine-tuning}). Ablations demonstrate that improvements are attributable to alignment of input–rationale–label triplets rather than to generic LM regularization; intentional misalignment leads to substantial drops in performance.
By \emph{moving reasoning to train time} and deploying a CoT-free classifier at test time, we show that DHRD preserves baseline latency while capturing much of CoT’s quality benefits for efficient reasoning.

\section{Related Work}
\label{sec:relatedwork}
\paragraph{Adapting decoder-only LMs for encoder-style classification.} Suganthan et al.\ introduce \emph{Gemma Encoder}, which adapts a decoder-only LM into an encoder by enabling bidirectional attention and adding task-specific pooling plus a Multilayer Perceptron (MLP) head; their analysis finds that simple pooling methods (last-token or mean) with an MLP head perform competitively with attention pooling. DHRD augments the pooling-MLP setup with a \emph{train-only} reasoning head so the model remains a pooled classifier at test time (no CoT generation). DHRD deliberately retains \emph{causal} attention and \emph{last-token} pooling to prevent future-token leakage and match autoregressive pretraining.

\paragraph{Relation to Rationale Supervision and Knowledge Distillation (KD).}
KD traditionally transfers knowledge from a teacher to a student model via soft-label or logit matching \citep{Hinton2015Distill}, and sequence-level KD is widely used in sequence-to-sequence settings \citep{KimRush2016SeqKD}. For LLMs, recent work explores distilling \emph{reasoning traces} or explanations (e.g., e-SNLI \citealt{Camburu2018eSNLI}). DHRD is a form of  \emph{reasoning-aware distillation}: we supervise the student with teacher-provided rationales, but \emph{only at train time}, and we do \emph{not} rely on answer logit matching or inference-time rationale generation. At test time, DHRD uses a pooled classification head (no CoT decoding).

\paragraph{Auxiliary Language-Modeling (LM) Loss.}
Adding auxiliary LM loss during supervised fine-tuning can improve generalization and accelerate convergence \citet{Radford2018Improving}. For our training, we use an auxiliary LM loss comprised of the input classification target plus rationale tokens produced by a CoT-capable teacher model. Our ablations indicate that the lift stems from \emph{aligned} input–rationale–label triplets rather than generic LM regularization.

\section{Method}
\label{sec:method}
\begin{figure}[htpb]
    \centering
    \includegraphics[width=1\linewidth]{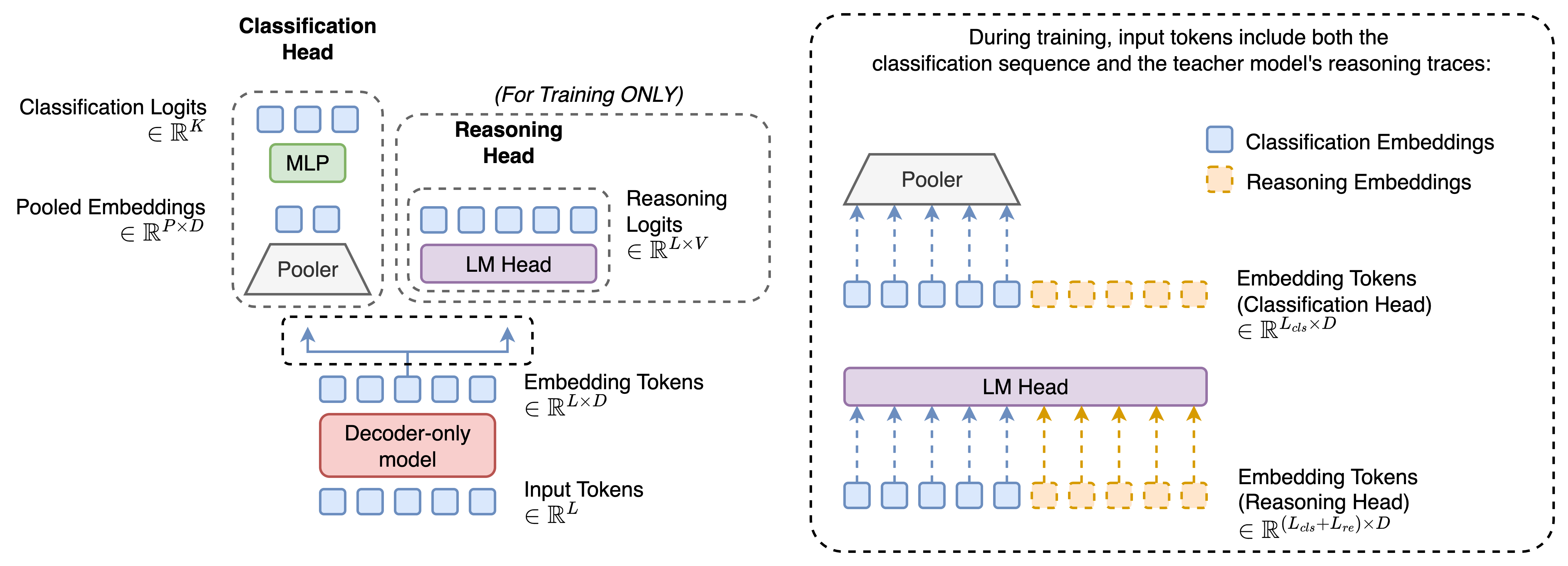}
    \caption{Dual-head fine-tuning on a shared decoder. The \emph{classification head} pools hidden states over the input span (blue) to produce $K$ class logits. The train-only \emph{reasoning head} applies a causal LM loss over the full sequence, covering both classification input tokens (blue) and teacher rationale tokens (orange). During training, inputs concatenate task text with teacher rationales. At inference, only the classification head is used.}
    \label{fig:placeholder}
\end{figure}

\subsection{Dual-Head Architecture}
We attach two lightweight heads to a shared decoder-only transformer with causal attention, enabling a single forward pass that supports both classification and train-time reasoning.

\paragraph{Training-time input and slicing} For a sample $i$, we form a single sequence by concatenating the original classification input $x_i$ with a teacher rationale $r_i$ (Gemini 2.5 Flash) and the gold label $y_i$:
\begin{equation}
\label{eq:train-seq}
\begin{aligned}
s_i &= [\,x_i,\ \text{\texttt{<REASON>}},\ r_i,\ \text{\texttt{<ANS>}},\ y_i\,],\qquad
L_i = |s_i|, \qquad L^{(x)}_i = |x_i|.
\end{aligned}
\end{equation}

\paragraph{Classification head.}
We pool the hidden state of the last real input token (before \texttt{<REASON>}) and map to $K$ logits (for $K$ classes) via a 2-layer Multi-layer Perceptron (MLP) following the adaptation proposed in ~\citet{Suganthan2025Adapting}. This is the only head used for prediction at test time.

\paragraph{Reasoning (LM) head, train-time only.}
We reuse the base model’s LM head to obtain next-token distributions; the standard causal LM loss is computed over the entire sequence $s_i$.

\subsection{Weighted Objective}

Let batch size be $B$, $\mathbf{z}_i\in\mathbb{R}^K$ be class logits for example $i$, and $y_i\in\{1,\dots,K\}$ the class label. The classification loss (explicit log-softmax form) is
\begin{equation}
\mathcal{L}_{\mathrm{cls}}
= -\frac{1}{B}\sum_{i=1}^B
\Big(
\mathbf{z}_i[y_i] - \log\!\sum_{k=1}^K e^{\mathbf{z}_i[k]}
\Big).
\end{equation}

For reasoning (causal LM), let the vocabulary size be $V$, batch size $B$, sequence lengths $L_i$,
token logits $\mathbf{\ell}_{i,t}\in\mathbb{R}^V$ at position $t$, and targets
$v_{i,t}\in\{1,\dots,V\}$ and a binary mask $m_{i,t}\in\{0,1\}$ indicating whether position $t$ is valid (so the loss uses $m_{i,t+1}$ to mask the next-token target). The reasoning loss is standard causal-LM (next-token) cross-entropy over the entire sequence:
\begin{equation}
\mathcal{L}_{\mathrm{reason}}
= -\,\frac{1}{N}\sum_{i=1}^{B}\sum_{t=1}^{L_i-1}
m_{i,t+1}\,
\log\!\left(
\frac{\exp\{\ell_{i,t}[\,v_{i,t+1}\,]\}}
{\sum_{w=1}^{V}\exp\{\ell_{i,t}[w]\}}
\right),
\qquad
N=\sum_{i=1}^{B}\sum_{t=1}^{L_i-1} m_{i,t+1}.
\end{equation}

The joint objective is
\begin{equation}
\mathcal{L}_{\mathrm{total}}
= \beta\,\mathcal{L}_{\mathrm{cls}} + \alpha\,\mathcal{L}_{\mathrm{reason}},
\quad \alpha,\beta \ge 0.
\end{equation}

At inference, we input only $x$, ignore the reasoning head, and produce $\mathbf{z}$ from pooled encoder states.


\section{Evaluation}
\label{sec:evaluation}

\subsection{Datasets}
We evaluate on seven SuperGLUE tasks \citep{SuperGLUE2019}: BoolQ \citep{Clark2019BoolQ}, CB \citep{DeMarneffe2019CB}, COPA \citep{Roemmele2011COPA}, MultiRC \citep{Khashabi2018MultiRC}, RTE \citep{Dagan2006RTE}, WiC \citep{Pilehvar2019WiC}, and WSC \citep{Levesque2012WSC}. ReCoRD is excluded as an extractive QA task. Following common practice, CB is reported as F1/Accuracy and MultiRC as F1a/EM; Avg is a macro-average across the seven tasks after first averaging the two metrics for CB and MultiRC. A brief summary of task formats and dataset sizes appears in Appendix~\ref{app:sg-summary}.
\begin{table}[htpb]
\centering
\caption{SuperGLUE results (higher is better). \textit{Rel.~$\Delta$~(\%)} is the relative percentage change versus the pooled baseline for the same backbone ($\alpha{=}0,\beta{=}1$). All DHRD rows use the optimal weights selected on the validation split: $\alpha{=}\beta{=}1$ for Llama-3.1-8B, Llama-3.2-3B, and Qwen-3-8B; $\alpha{=}0.5,\ \beta{=}1$ for Qwen-3-4B. Full $\alpha/\beta$ ablations can be found in Appendix~\ref{app:loss-weight}.}

\label{tab:sg-eval}
\small                      
\setlength{\tabcolsep}{3pt} 
\begin{adjustbox}{max width=\columnwidth} 
\begin{tabular}{l l c c c c c c c c c}
\toprule
Model & Setting &BoolQ & CB & COPA  & MultiRC & RTE & WiC & WSC & {Avg} & {Rel. $\Delta$ (\%)}\\
\midrule
Gemini 2.5 Flash & CoT Zero-shot & 88.8 &  83.1/92.0 & 98.4  & 86.2/59.1 & 91.3 & 71.6 & 94.5 & 86.40 \\
\midrule Llama-3.1-8B & Baseline (pooled classifier)  & 90.7 &  81.9/90 & 93.8  & 88.9/62.9 & 91.4 &\textbf{ 74.8} & \textbf{91.5} & 86.29 & {} \\

             & DHRD (train-time reasoning)  & \textbf{91.1} &\textbf{ 89.0/94.8 }& \textbf{95.4} & \textbf{89.0/63.7} & \textbf{92.2 }& 74.6 & 91.1 & \textbf{87.52}  & \bfseries+1.43\%
              \\
   
\midrule
Qwen-3-8B     
& Baseline (pooled classifier)  &89.4 & 83.5/93.2 &  \textbf{93.8} & 88.2/62.0  & 91.9 & 75.1 & 89.0 & 86.09 & {}  \\
      
             & DHRD (train-time reasoning) & \textbf{90.5}

&\textbf{90.8/95.6} & 93.2 & \textbf{88.5/62.5} & \textbf{91.9} & \textbf{76.6 }& \textbf{89.7} & \textbf{87.23} & \bfseries +1.32\% \\
 
 \midrule
Llama-3.2-3B & Baseline (pooled classifier) & 89.0 & \textbf{78.6/88.8} &  72.2 & 83.5/49.8  & 84.3 & 68.1 & 77.4 & 77.34& {} \\
        
             & DHRD (train-time reasoning) &\textbf{89.2} & 73.8/87.2& \textbf{89.2} & \textbf{85.9/55.3} & \textbf{87.7} & \textbf{73.0} & \textbf{80.8} & \textbf{81.57} &\bfseries +5.47\% \\
      
             \midrule
Qwen-3-4B     & Baseline (pooled classifier) &88.9 & 84.3/91.6 &  92.4 & 87.7/62.0  & 90.4 & 71.4 & 85.6 & 84.50  & {} \\
             & DHRD (train-time reasoning) & \textbf{89.5} & \textbf{84.4/92.0} &\textbf{92.8} &\textbf{ 88.1/62.4} & \textbf{90.8} & \textbf{71.4} & \textbf{87.4} & \textbf{85.05} & \bfseries +0.65\% \\

\bottomrule
\end{tabular}

\end{adjustbox}
\end{table}

\paragraph{SuperGLUE Results.} 
 Table~\ref{tab:sg-eval} shows that the 8B models perform best with ($\beta{=}\alpha{=}1$); Llama-3.2-3B sees the largest relative lift ($+5.47\%$), while Qwen3-4B prefers a lower $\alpha{=}0.5$. Both 8B DHRD settings exceed the teacher model's (Gemini 2.5 Flash) average SuperGLUE score.

  Train-time reasoning improves accuracy most on \emph{entailment and cause–effect} style tasks (CB, RTE, COPA), both Llama and Qwen 8B improved by +8.7\% accuracy on CB, and Llama 3B improved by +23.5\% accuracy on COPA; consistent improvements are observed in RTE, MultiRC, and BoolQ.  WiC/WSC are relatively stable, likely because word-sense and coreference tasks are less reliant on multi-step verbalization.

\subsection{Inference Efficiency (no CoT vs. CoT)}
Since DHRD disables rationale generation at test time, the relevant throughput is the pooled \emph{classification} head’s forward pass (no decoding). Appendix ~\ref{app:qps} shows that DHRD preserves pooled-classifier throughput and \emph{avoids} CoT decoding costs. For context, we also report the throughput of CoT-style decoding on the same backbones (for settings refer to Appendix~\ref{app:reasoning-head-CoT}), which shows that the pooled path is (96--142$\times$) faster than CoT-style inference on the same decoder-only backbones. 

\subsection{Ablations: Reasoning}\label{app:ablations}
\begin{table}[htpb]
\centering
\caption{Ablations on rationale/label alignment (SuperGLUE). ConsistentReasoningLabel (aligned \texttt{<REASON>} and \texttt{<ANS>}), OnlyLabel (aligned \texttt{<ANS>}), ShuffleReasoning (misaligned \texttt{<REASON>}, aligned \texttt{<ANS>}), ShuffleReasoningLabel (misaligned \texttt{<REASON>} and \texttt{<ANS>}). In all settings, the LM loss is applied to the input tokens and present \texttt{<REASON>}/\texttt{<ANS>} segments. All rows use $\alpha{=}\beta{=}1$.}

\label{tab:ablation}\small                      
\setlength{\tabcolsep}{3pt} 
\begin{adjustbox}{max width=\columnwidth} 
\begin{tabular}{l l c c c c c c c c c}
\toprule
Model & DHRD Setting &BoolQ & CB & COPA  & MultiRC & RTE & WiC & WSC & {Avg} & {Rel. $\Delta$ (\%)} \\

\midrule
\addlinespace[2pt]
Llama-3.1-8B & ConsistentReasoningLabel & \textbf{91.1} & \textbf{89.0/94.8} & \textbf{95.4} & \textbf{89.0/63.7} & \textbf{92.2} & 74.6 &\textbf{ 91.1 }&\bfseries 87.52 & {} \\
             & OnlyLabel &  90.7 & 87.6/93.6  & 93.8 & 84.6/53.2 &  91.1 & \textbf{76.4} &  90.4 & 85.99 & -1.75\% \\
              & ShuffleReasoning & 90.4 & 59.5/76.4 & 95.2 & 84.9/53.1 & 90.9 & 75.9 & 88.4 & 82.54 & -5.70\% \\
             & ShuffleReasoningLabel &62.2 & 35.4/51.6 & 44.6 & 0.0/0.5 & 50.0 & 50.0 & 65.1 & 45.09& -48.48\% \\
\midrule
Llama-3.2-3B & ConsistentReasoningLabel & \textbf{89.2} & \textbf{73.8/87.2} & 89.2 & \textbf{85.9/55.3} & \textbf{87.7 }& 73.0 & \textbf{80.8} & \textbf{81.57}& {} \\
             & OnlyLabel &   88.0  & 54.5/79.2 & \textbf{91.4} & 84.5/51.3 &  84.3 &\textbf{ 75.3 }   & 75.34 & 78.44 & -3.84\% \\
              & ShuffleReasoning & 81.4 & 57.8/76.0 & 90.2  & 11.2/0.6 & 87.3 & 68.6  & 55.12 & 65.06 & -20.24\% \\
             & ShuffleReasoningLabel &62.3 & 21.7/48.4 & 50.0 & 0.0/0.5 & 50.1 & 49.7 & 65.1 & 44.64 & -45.27\%\\
\bottomrule
\end{tabular}
\end{adjustbox}

\end{table}
\paragraph{Ablation Results.}
Table~\ref{tab:ablation}  compares our best setting (ConsistentReasoningLabel) to three controls that remove or corrupt the explanation signal.
The results show removing rationales hurts (OnlyLabel), while misalignment degrades performance, especially on entailment/causal tasks (CB, RTE, COPA).

\section{Conclusion}
\label{sec:conclusion}
\emph{Dual-Head Reasoning Distillation} (DHRD) is a reasoning-classification joint-training technique that moves the computational cost of reasoning to training time, improving accuracy without inference overhead. Across seven SuperGLUE tasks and 3–8B decoder-only backbones, DHRD improves pooled baselines by \textbf{0.65–5.47\%} (macro-avg) with the largest gains on entailment/causal tasks (CB, RTE, COPA). With no rationale generation at inference time, DHRD preserves the pooled classifier’s latency and achieves \textbf{96–142$\times$} higher throughput than CoT-style decoding on the same backbones. Ablations show the lift comes from \emph{aligned} input–rationale–label triplets rather than generic LM regularization: removing or misaligning rationales degrades accuracy, especially on entailment. In practice, we find $\alpha{=}1$ works well for 8B models while Qwen 4B model prefer a milder LM weight ($\alpha{\approx}0.5$). Limitations include dependence on the quality of rationale and sensitivity to the LM-loss weight; we also report single runs per setting due to computational constraints (refer to Appendix ~\ref{app:broader-limitations}).

{\small
\nocite{Suganthan2025Adapting,SuperGLUE2019,Qwen3TechReport2025,Llama3Herd2024,Radford2018Improving,Wei2022CoT,Hinton2015Distill,KimRush2016SeqKD,Wolf2020Transformers,
Clark2019BoolQ,DeMarneffe2019CB,Roemmele2011COPA,Khashabi2018MultiRC,Dagan2006RTE,Pilehvar2019WiC,Levesque2012WSC,Gemini25Report2025,Bach2022PromptSource,Khashabi2020UnifiedQA,LeScaoRush2021PromptWorth,Wang2022SelfConsistency,Zhou2022LeastToMost,
Yao2023TreeOfThoughts,Wang2023PlanAndSolve,
Vaswani2017Attention,Devlin2019BERT,Brown2020GPT3,Kaplan2020ScalingLaws,
Hoffmann2022Chinchilla,
Snell2024TestTimeCompute,Li2024ESC,Camburu2018eSNLI,Zhang2021ExPred,Kumar2020NILE,HadjiKyriacou2024DPH,Yang2024DualHeadKD,
Shridhar2023Distilling,Chen2023MCCKD,Carton2022Rationales,Zhu2025RationalesNotSilverBullets,Nair2010ReLU,Hendrycks2016GELU,Rumelhart1986Backprop, Liu2024HiddenCoT, Kang2024C3oT, Cheng2024CCoT,Kojima2022ZeroShotNeurIPS, Shukla2023UGCEmbeddings, Hu2022LoRA}}

\bibliographystyle{plainnat} 
\bibliography{refs}

\newpage

\appendix
\section{Broader Impact and Limitations}
\label{app:broader-limitations}

\textbf{Broader impact.} \emph{Dual-Head Reasoning Distillation} (DHRD) retains the accuracy benefits of chain-of-thought supervision while restoring pooled-classifier throughput by removing inference-time rationales, lowering latency and compute. We observe the largest gains on entailment and cause–and–effect tasks, which can strengthen knowledge-intensive QA by verifying whether evidence \emph{supports} or \emph{contradicts} an answer and improve safety moderation by checking semantic \emph{entailment} of policy violations rather than keywords. 

Risks include reduced transparency and contestability when rationales are not emitted and amplification of biases under distribution shift. There is also a possibility of false positives/negatives in moderation or QA, potential misuse for high-throughput surveillance or censorship, and rebound effects from scaling automated decisions. To mitigate these, we commit to detailed model cards with per-group performance and calibration with human-in-the-loop review and appeal mechanisms in high-stakes settings. We avoid training on sensitive personal information and consider staged/gated release with usage policies.

\textbf{Limitations.}
Observed gains depend on high-quality, label-consistent teacher rationales; removing or misaligning rationales substantially harms accuracy, particularly on entailment/causal tasks. Performance is sensitive to the auxiliary LM-loss weight $\alpha$: on 4B backbones, large $\alpha$ can regress accuracy (moderate values, e.g., $\alpha{\approx}0.5$, are safer), whereas 8B models tolerated $\alpha{=}1$; more systematic schedules are left to future work. Our evaluation covers seven SuperGLUE tasks (excluding ReCoRD) and 3--8B decoder-only backbones with last-token pooling; generalization to other languages, modalities, domains, and architectures remains untested.

We have not yet conducted a direct, budget-controlled comparison to alternative \emph{CoT compression} methods such as HiddenCoT, C3oT, and CCoT, which target reduced rationale tokens at training and/or inference \citep{Liu2024HiddenCoT,Kang2024C3oT,Cheng2024CCoT}. As a result, it remains unclear how DHRD trades off accuracy, calibration, and latency relative to these approaches. 

Due to compute constraints, we report single runs without confidence intervals; we partially mitigate this by reporting per-task metrics and macro-averages, showing trends across models and $\alpha$ values, and specifying exact training/eval configurations to aid replication (App.~\ref{app:hparams}).

\section{Loss Weighting Ablations} \label{app:loss-weight}
\begin{table}[htpb]
\centering
\caption{SuperGLUE results (higher is better). {Rel. $\Delta$ (\%)} is the relative percentage change compared to each model's pooled baseline ($\alpha{=}0,\beta{=}1$). Reasoning / CoT fine-tuned models follows ($\alpha{=}1,\beta{=}0$) with CoT at inference using the Reasoning Head (refer to Appendix~\ref{app:reasoning-head-CoT}).}

\label{tab:sg-main}
\small                      
\setlength{\tabcolsep}{3pt} 
\begin{adjustbox}{max width=\columnwidth} 
\begin{tabular}{l l c c c c c c c c c}
\toprule
Model & Setting &BoolQ & CB & COPA  & MultiRC & RTE & WiC & WSC & {Avg} & {Rel. $\Delta$ (\%)}\\
\midrule
Gemini 2.5 Flash & CoT Zero-shot & 88.8 &  83.1/92.0 & 98.4  & 86.2/59.1 & 91.3 & 71.6 & 94.5 & 86.40 \\
\midrule
Llama-3.1-8B & Reasoning / CoT  & 80.6 &53.5/77.6 & 91.8  & 71.5/27.4 &71.7 & 66.3 & 76.12 & 71.65\\
\midrule
Qwen3-8B     
&  Reasoning / CoT  & 81.8 & 63.5/70.4 & 92.6  &81.4/44.7 & 90.9 & 74.3  & 77.4 & 78.14  \\
\midrule Llama-3.1-8B & Baseline & 90.7 &  81.9/90 & 93.8  & 88.9/62.9 & 91.4 & 74.8 & 91.5 & 86.29 & {} \\
 
             & DHRD $(\beta{=}1,\alpha{=}0.5)$ &
             91.0 & \textbf{89.1/94.8} & 94.8 & 88.5/63.3 & \textbf{92.3} & 72.5 & \textbf{92.5} & 87.28 & +1.15\%\\
             
             & DHRD $(\beta{=}1, \alpha{=}1)$ & \textbf{91.1} & 89.0/94.8 & \textbf{95.4} & 89.0/63.7 & 92.2 & 74.6 & 91.1 & \textbf{87.52}  & \bfseries+1.43\%
              \\
             & DHRD $(\beta{=}0.5, \alpha{=}1)$ &  90.7 & 85.6/92.8 & 94.4 & \textbf{89.1/63.8} &  91.1 & \textbf{74.8} & 88.4 & 86.44 & +0.17\%\\
\midrule
Qwen3-8B     
& Baseline &89.4 & 83.5/93.2 &  93.8 & 88.2/62.0  & 91.9 & 75.1 & 89.0 & 86.09 & {}  \\
             & DHRD $(\beta{=}1,\alpha{=}0.5)$ & 89.7 & \textbf{91.8/95.6 }&92.8 & 88.2/62.0 &91.8 & 76.4 & 88.4 & 86.84 & +0.87\% \\
             & DHRD $(\beta{=}1, \alpha{=}1)$ & 90.5

&90.8/95.6 & 93.2 & \textbf{88.5/62.5} & \textbf{91.9} & 76.6 & \textbf{89.7} & \textbf{87.23} & \bfseries +1.32\% \\
             &  DHRD $(\beta{=}0.5, \alpha{=}1)$ &  \textbf{90.6} & 89.3/95.2
 & \textbf{94.0} & 87.5/60.1 &91.7 & \textbf{77.9} & 87.7 & 86.85 & +0.88\% \\ 
 \midrule
Llama-3.2-3B & Baseline & 89.0 & \textbf{78.6/88.8} &  72.2 & 83.5/49.8  & 84.3 & 68.1 & 77.4 & 77.34& {} \\
             & DHRD $(\beta{=}1,\alpha{=}0.5)$ &  89.1 & 52.4/76.4 & 89.6 & 86.4/57.7 & \textbf{87.8} & 72.5 & \textbf{84.2} & 79.95 & +3.37\% \\
             & DHRD $(\beta{=}1, \alpha{=}1)$ &\textbf{89.2} & 73.8/87.2& 89.2 & 85.9/55.3 & 87.7 & \textbf{73.0} & 80.8 & \textbf{81.57} &\bfseries +5.47\% \\
             & DHRD $(\beta{=}0.5, \alpha{=}1)$ & 88.3 & 62.6/78.4 & \textbf{90.8} & \textbf{87.2/59.1} & 86.8 & 56.6 & 77.4 & 77.65 & +0.40\% \\
             \midrule
Qwen3-4B     & Baseline &88.9 & 84.3/91.6 &  92.4 & 87.7/62.0  & 90.4 & 71.4 & 85.6 & 84.50  & {} \\
             & DHRD $(\beta{=}1,\alpha{=}0.5)$ & 89.5 & \textbf{84.4/92.0} &92.8 &\textbf{ 88.1/62.4} & 90.8 & 71.4 & \textbf{87.4} & \textbf{85.05} & \bfseries +0.65\% \\
             & DHRD $(\beta{=}1, \alpha{=}1)$ &\textbf{89.6}
& 73.0/82.4 & \textbf{93.2} & 87.9/60.5 & \textbf{91.2} & 71.8 & 87.6 & 83.61 & -1.05\% \\
             & DHRD $(\beta{=}0.5, \alpha{=}1)$ &88.5 & 68.2/88.0 & 91.4 & 87.5/60.1 & 90.1 & \textbf{72.6} & 84.2 & 82.67  & -2.17\% \\
\bottomrule

\end{tabular}
\end{adjustbox}

\end{table}

\newpage
\section{SuperGLUE Dataset Overview}\label{app:sg-summary}

\begin{table}[htpb]
\centering
\caption{SuperGLUE benchmark overview with task type, train, validation, and test dataset sizes}
\label{tab:sg-summary}
\small                      
\setlength{\tabcolsep}{3pt} 
\begin{adjustbox}{max width=\columnwidth}
    \begin{tabular}{lccccc}
        \toprule
        Benchmark (SuperGLUE) & Task & Metric & Train  & Validation & Test \\ \midrule
        BoolQ & QA & Accuracy & 9427  & 3270 & 3245 \\
        CB & NLI & F1/Accuracy & 250 & 56 & 250   \\
        COPA & QA & Accuracy & 400 & 100 & 500  \\
        MultiRC & QA & F1/EM & 27243 & 4848 & 9693  \\
        RTE & NLI & Accuracy & 2490 & 277 & 3000   \\
        WiC & WSD & Accuracy & 5428 & 638 & 1400  \\
        WSC & Coref. & Accuracy & 554 & 104 & 146  \\
        \bottomrule
\end{tabular}
\end{adjustbox}
\end{table}
\section{QPS Results on same decoder backbones (no CoT vs. CoT)}\label{app:qps}
\begin{table}[htpb]
\centering
\caption{Throughput (queries per second, higher is better). Classification uses a pooled head at inference (no decoding). Reasoning uses CoT-style decoding (train-time only in DHRD). The rightmost column shows the speedup of our deployed path over CoT decoding on the same backbone.}
\label{tab:throughput-table}
\small
\setlength{\tabcolsep}{6pt}
\begin{tabular}{lccc}
\toprule
Model & QPS (Pooled Classifer) & QPS (Reasoning / CoT) & Speedup (no-CoT / CoT) \\
\midrule
Llama-3.1-8B & 414.91 & 4.14  & $\sim\!100\times$ \\
Llama-3.2-3B & 631.16 & 4.44  & $\sim\!142\times$ \\
Qwen3-8B     & 341.15 & 3.56  & $\sim\!96\times$ \\
Qwen3-4B     & 440.83 & 3.76  & $\sim\!117\times$ \\
\bottomrule
\end{tabular}
\end{table}

\section{Fine-tuning Details}\label{app:fine-tuning}

\subsection{Hardware}\label{app:hardware}
All experiments were run with distributed data parallel training on 8$\times$ H100 (80GB) GPUs using NCCL. Unless noted otherwise, per-device train/eval batch size was 1 with gradient accumulation to reach effective batch sizes (see \S\ref{app:hparams}). Mixed precision used \texttt{bf16}.

\subsection{Prompt Templates (UnifiedQA-style)}
\label{app:prompts}
We adopt a single \emph{UnifiedQA-style} prompt across tasks \citep{Khashabi2020UnifiedQA}, phrased to normalize inputs to a QA format. The student’s \textbf{classification input} is:

\begin{quote}\small
\texttt{Passage: \{paragraph\}\\
Question: \{question\}\\
Answer: \{answer\}\\
Is the answer correct Yes or No?}
\end{quote}

During training, we append a train-time only \textbf{reasoning suffix} (teacher rationale $r$ and gold label $y$) so the reasoning head can compute token-level likelihood:
\begin{quote}\small
\texttt{Reasoning: \{explanation\}\\
Final Answer: \{label\}}
\end{quote}

At inference, only the classification head is used; no rationale is generated.

\paragraph{Task mapping.} We convert each SuperGLUE task to the same schema:
\begin{itemize}\itemsep0.2em
  \item \textbf{MultiRC}: \texttt{paragraph}$\to$ passage, \texttt{question}$\to$ question, candidate \texttt{answer}$\to$ answer; label is Yes/No for correctness.
  \item \textbf{BoolQ}: passage is the Wikipedia paragraph; the Boolean \texttt{answer} becomes the candidate; label is whether “Yes” is correct.
  \item \textbf{CB/RTE}: cast as “Is the hypothesis entailed by the premise? \texttt{Answer}: "Yes/No”.
  \item \textbf{COPA}: premise as passage; candidate cause/effect as \texttt{answer}; label indicates the correct alternative (Yes if the shown choice is correct).
  \item \textbf{WiC}: sentence pair as \texttt{Passage} with the target word highlighted; \texttt{Question}: “Does the word have the same meaning?”; \texttt{Answer}: “Yes/No”.
  \item \textbf{WSC}: passage contains the pronoun and candidate antecedent; \texttt{Question}: “Does the pronoun refer to \{candidate\}?”; \texttt{Answer}: “Yes/No”.
\end{itemize}
This unified phrasing stabilizes the pooled classifier and makes the rationale generator consistent across tasks, following the spirit of UnifiedQA’s “single prompt for many formats.”

\subsection{Teacher Prompt for Rationale Generation}

\label{app:teacher}
Rationales $r$ are produced by a CoT-capable teacher (Gemini 2.5 Flash v3; \citealp{Gemini25Report2025})  using a guarded prompt. The gold label is provided to the teacher but \emph{the rationale is not allowed to restate the label}. We enforce a sentinel to separate explanation and answer.

\begin{quote}\small
\textbf{Instruction (to teacher).} Given the passage, question, and candidate answer, write a short explanation (2--5 sentences) that justifies the \emph{gold label}. Let's think step by step. \\
\textbf{Rules:} (1) Do \emph{not} include “Yes/No”, “True/False”, or synonyms in the explanation; (2) End the explanation with \texttt{[END OF REASONING]}; (3) The explanation should be a short paragraph. (3) On a new line, output \texttt{Answer: Yes} or \texttt{Answer: No}.\\[0.25em]
\textbf{I/O Format}\\
\texttt{Passage: ...\\
Question: ...\\
Answer: ...\\
Reasoning: <concise explanation not revealing the label> [END OF REASONING]\\
Answer: <Yes|No>}
\end{quote}

\subsection{LoRA Configuration}
\label{app:lora}
We adapt the backbone with PEFT-LoRA applied to attention and MLP projections:
\begin{center}\small
\begin{tabular}{ll}
\toprule
Target modules & \texttt{[q\_proj, k\_proj, v\_proj, o\_proj, up\_proj, down\_proj, gate\_proj]}\\
Rank $r$ / $\alpha$ / dropout & \textbf{$r=16$}, \textbf{$\alpha=32$}, dropout $=0.1$ \\
Bias & \texttt{none} \\
Task type & \texttt{CAUSAL\_LM} \\
\bottomrule
\end{tabular}
\end{center}
Only the LoRA adapters and the small classification head are trainable.

\subsection{Training Hyperparameters}
\label{app:hparams}
We keep the base decoder architecture unchanged and optimize a weighted loss $\mathcal{L}_\text{total}=\beta\mathcal{L}_\text{cls}+\alpha\mathcal{L}_\text{reason}$.
\begin{itemize}\itemsep0.2em
  \item \textbf{Optimizer}: AdamW (\texttt{optim=adamw\_torch}), LR $2\!\times\!10^{-4}$, weight decay $0.01$.
  \item \textbf{Schedule}: cosine with \texttt{warmup\_steps=5}.
  \item \textbf{Precision}: \texttt{bf16}.
  \item \textbf{Batching}: per-device batch size $=1$; \texttt{gradient\_accumulation\_steps=32}.
  \item \textbf{Epochs}: $3$; \texttt{save\_strategy=eval\_strategy=epoch}.
  \item \textbf{Loss weights}: unless otherwise specified in tables, we report \texttt{Reasoning1} with $\beta{=}1,\alpha{=}1$ (and ablate $\alpha\in\{0.5,1.0\}$).
  \item \textbf{Collator}: dual-sequence collator that pads classification and explanation streams independently and masks explanation padding to \texttt{-100}.
  \item \textbf{DDP}: \texttt{ddp\_backend=NCCL}, \texttt{ddp\_find\_unused\_parameters=True}.
\end{itemize}

\subsection{Generation Settings for Reasoning Head CoT} \label{app:reasoning-head-CoT}
When we generate rationales (or analyze decoding throughput), we use:
\begin{itemize}\itemsep0.2em
  \item \texttt{do\_sample=True}, \texttt{temperature=0.1}, \texttt{top\_p=0.7}, \texttt{max\_new\_tokens=500}.
  \item EOS/pad from the tokenizer; sequences are decoded with special tokens skipped.
\end{itemize}

\subsection{Throughput (QPS) Measurement Protocol}
Classification-head QPS is computed as \emph{number of samples / wall-clock} for a pure forward pass (no text generation). Reasoning-head QPS measures decoding with the settings above. For multi-GPU, we report per-rank stats and an aggregated global QPS using a max-reduce of wall-clock time across ranks to avoid optimistic scaling.

\section{Future Work}
We plan to explore larger backbones and grid search for optimal $\alpha,\beta$ values and introduce task-aware auxiliary objectives (e.g., span supervision for WiC/WSC). We also plan to investigate curriculum schedules that gradually anneal the reasoning loss and explore knowledge distillation with soft labels for the train-time reasoning head (e.g., temperature-scaled KL divergence between teacher and student token distributions over the input{+}rationale span) to test whether soft targets improve stability and transfer without changing inference cost. We also should consider running head-to-head evaluations against recent \emph{chain-of-thought compression} approaches (e.g., HiddenCoT, C3oT, and CCoT) under matched token budgets and latency constraints, reporting accuracy-per-token and robustness to teacher-rationale quality \citep{Liu2024HiddenCoT,Kang2024C3oT,Cheng2024CCoT}. We will also explore hybrid variants that combine DHRD with lightweight rationale selection or summary-style sketches to further reduce reasoning tokens while preserving accuracy.

\section{Assets \& Licenses}
\label{app:assets-licenses}

We use third-party datasets and models under their original terms; all use is for non-commercial research. Table~\ref{tab:assets-licenses} lists the assets and licenses.

\begin{table}[htbp]
\centering
\caption{Third-party assets and licenses.}
\label{tab:assets-licenses}
\small
\setlength{\tabcolsep}{6pt}
\begin{adjustbox}{max width=\linewidth}
\begin{tabular}{lllp{7cm}}
\toprule
\textbf{Asset} & \textbf{Type} & \textbf{License} & \textbf{Notes}\\
\midrule
BoolQ \citep{Clark2019BoolQ} & Dataset & CC BY-SA 3.0 & Used via SuperGLUE splits for research.\\
CB \citep{DeMarneffe2019CB} & Dataset & Original terms & Research use per SuperGLUE inclusion; cite original source.\\
COPA \citep{Roemmele2011COPA} & Dataset & Original terms & Research use per SuperGLUE inclusion.\\
MultiRC \citep{Khashabi2018MultiRC} & Dataset & Original terms & Research use per SuperGLUE inclusion.\\
RTE \citep{Dagan2006RTE} & Dataset & Original terms & Research use per SuperGLUE inclusion.\\
WiC \citep{Pilehvar2019WiC} & Dataset & CC BY-NC 4.0 & Non-commercial.\\
WSC \citep{Levesque2012WSC} & Dataset & CC BY 4.0 & —\\
Llama 3.x \citep{Llama3Herd2024} & Model & Meta Llama Community License & Used for research fine-tuning.\\
Qwen 3 \citep{Qwen3TechReport2025} & Model & Apache-2.0 & —\\
Gemini 2.5 Flash \citep{Gemini25Report2025} & API outputs & Gemini API Additional Terms & Used to generate training rationales.\\
\bottomrule
\end{tabular}
\end{adjustbox}
\end{table}

\textbf{Compliance.} We follow the license obligations for each asset (attribution, non-commercial restrictions, and terms of service) and use these assets solely for research. We do not redistribute datasets or model weights.

\end{document}